\documentclass[twoside,11pt]{article}

\usepackage{jmlr2e}
\usepackage{mathtools}
\usepackage{graphicx}
\usepackage{epsfig}
\usepackage{subfigure}

\newcommand{\dataset}{{\cal D}}

\ShortHeadings{Optimal Linear Combination of Classifiers}{Nalbantov and Ivanov}
\firstpageno{1}

\begin{document}

\title{Optimal Linear Combination of Classifiers}

\author{\name Georgi Nalbantov \email gnalbantov@yahoo.com \\
	\name Svetoslav Ivanov \email svetoslav.val.ivanov@gmail.com \\
       \addr Department of Data Science, Data Science Consulting Ltd., Bulgaria}

\editor{}
\today

\maketitle

\begin{abstract}
The question of whether to use one classifier or a combination of classifiers is a central topic in Machine Learning.  We propose here a method for finding an optimal linear combination of classifiers derived from a bias-variance framework for the classification task.
\end{abstract}


\section{Introduction}

Numerous approaches exist on ways to combine the output of classifiers to produce a final prediction. We present here an approach within the bias-variance framework for the classification task to combine the output of classifiers linearly so as to produce a better out-of-sample prediction. We consider the binary classification task.

\section{Bias-variance framework for classification}
\label{sec:BV}

The bias-variance framework is applied to judge the theoretical performance of a regression or classification model. Within this framework the expected performance in terms of expected prediction error is split into three components, so-called bias, variance, and irreducible (Bayes) error. In case of regression, the components form a linear combination, whereas in the case of classification there are interaction terms as well. We first derive the bias-variance decomposition for single (binary) classification models before putting classifier combinations into this framework. 

The expected prediction error in case of classification is defined in terms of the 0-1 loss function: a loss of 1 is incurred in case of misclassification, and 0 otherwise. In a binary-target ($\in \{0,1\}$) dataset $\dataset$ we first consider the bias-variance decomposition at a specific point $\bf{x}$ (rather than over the whole domain of the dataset). Theoretically, the performance of any classifier is estimated over an infinite number of datasets generated by the underlying function which has generated dataset $\dataset$ (for a given fixed number of points in a dataset). The expected loss at point $\bf{x}$ over these datasets $\dataset$ is the proportion of time the classifier outputs incorrect prediction (at this point), which is equal to the probability of misclassification:

\begin{equation}
\label{BVdecomp}
E_{\dataset,t} \left( L\left(  t,y  \right)   \right) = P(t \neq y),
\end{equation}

\noindent where $E$ denotes expectation,  $L$ is the 0-1 loss function, $t \in \{0,1\}$ is realized class of point $\bf{x}$ in a dataset $\dataset$,  $P$ is probability taken over realizations of $t$ over different training sets $\dataset$, and $y \in \{0,1\}$ is a (fixed) predicted class at $\bf{x}$ in a test dataset $\dataset$. The classifier with best theoretical performance at point $\bf{x}$ is the one which prediction is the class which occurs with higher frequency within the (infinite amount of) datasets. This classifier is referred to as the Bayes classifier. We denote the prediction of the Bayes classifier, which is the theoretically optimal prediction, at a point $\bf{x}$ as $y^*$, where $y^* \in \{0,1\}$. Therefore, the predicted class $y$ of a given (non-Bayes) classifier should optimally be equal to $y^{*}$, that is $y$ is a prediction for $y^*$, denoted as $y = \widehat{y^*}$. We can expand Eq.(\ref{BVdecomp}) in the following way:

\[
E_{\dataset,t} \left( L\left(  t,y  \right)   \right) = P(t \neq y) = P( t\neq y^*  \cap   y = y^*) + P( t= y^*  \cap   y \neq y^*) =
\]
\begin{equation}
\label{BVdecomp_long}
P(t \neq y^{*} \mid y=y^{*}) P(y=y^{*}) +  P(t = y^{*} \mid y \neq y^{*}) P(y \neq y^{*})
\end{equation}

Under independence between $t$ and $y$, which occurs when $y$ is estimated from a training set, and $t$ is from a test set, Eq.(\ref{BVdecomp_long}) becomes:

\[
E_{\dataset,t} \left( L\left(  t,y  \right)   \right) = P(t \neq y^{*}) P(y=y^{*}) +  P(t = y^{*}) P(y \neq y^{*}) =
\]
\[
\text{BE} \left( 1- P(y \neq y^{*}) \right) + \left( 1 - \text{BE} \right) P(y \neq y^{*}) = 
\]
\[
\text{BE}  - \text{BE} \times  P(y \neq y^{*}) +  P(y \neq y^{*}) - \text{BE} \times  P(y \neq y^{*}) = 
\]
\[
P( y \neq y^{*}) \times (1 - 2 \text{BE}) + \text{BE} = 
\]
\begin{equation}
\label{BVdecomp_ind}
E_{\dataset,t} \left( L\left(  y,y^{*}  \right)   \right) \times (1 - 2 \text{BE})  + \text{BE},
\end{equation}

\noindent where we have defined the Bayes error as 
\[
\text{BE} \equiv E_{\dataset,t} \left( L\left(  t,y^{*}  \right)   \right) = P(t \neq y^{*}).
\]

If we denote the expected prediction at point $\bf{x}$ obtained from predictions made on each training dataset as $y^{m}$, that is $y^{m} = E_{\dataset,t}(y)$, then the expected loss between the classifier's prediction and the Bayes prediction at point $\bf{x}$, $E_{\dataset,t} \left( L\left(  y,y^{*}  \right)   \right)$, can be expanded as:

\[
E_{\dataset,t} \left( L\left(  y,y^{*}  \right)   \right) = P(y \neq y^{*}) = P( y = y^m  \cap   y^* \neq y^m) + P( y \neq y^m  \cap   y^* = y^m) = 
\]
\begin{equation}
\label{BVdecomp_yystar}
P( y = y^m  \mid   y^* \neq y^m) P(y^* \neq y^m) + P( y \neq y^m  \mid   y^* = y^m) P( y^* = y^m). 
\end{equation}

\noindent Under independence of $y$ and $y^*$, which is the case as the prediction $y^*$ of the Bayes classifier is not available to the classifier which prediction is $y$, Eq.(\ref{BVdecomp_yystar}) becomes:

\[
E_{\dataset,t} \left( L\left(  y,y^{*}  \right)   \right) = P( y = y^m ) P(y^* \neq y^m) + P( y \neq y^m ) P( y^* = y^m) =
\]
\[
 P( y = y^m ) P(y^* \neq y^m) + P( y \neq y^m ) (1 - P( y^* \neq y^m) ) =
\]
\[
 P( y = y^m ) P(y^* \neq y^m) + P( y \neq y^m )  - P( y \neq y^m )P( y^* \neq y^m) =
\]
\[
 P( y^* \neq y^m ) \left[ P( y = y^m ) - P( y \neq y^m ) \right] + P( y \neq y^m ) = 
\]
\[
 P( y^* \neq y^m ) \left[ 1 - 2 P( y \neq y^m ) \right] + P( y \neq y^m ) = 
\]
\[
 P( y^* \neq y^m ) - 2 P( y \neq y^m ) P( y^* \neq y^m ) + P( y \neq y^m ) = 
\]
\[
 P( y \neq y^m ) \left[ 1 - 2 P( y^* \neq y^m ) \right] + P( y^* \neq y^m ).
\]

That is:
\begin{equation}
\label{BVdecomp_yystar_short}
E_{\dataset,t} \left( L\left(  y,y^{*}  \right)   \right) = P( y \neq y^m ) \left[ 1 - 2 P( y^* \neq y^m ) \right] + P( y^* \neq y^m ).
\end{equation}

Plugging in Eq.(\ref{BVdecomp_yystar_short}) into Eq.(\ref{BVdecomp_ind}), we arrive at the bias-variance decomposition:

\[
E_{\dataset,t} \left( L\left(  t,y  \right)   \right) = E_{\dataset,t} \left( L\left(  y,y^{*}  \right)   \right) \times (1 - 2 \text{BE})  + \text{BE} = 
\]
\[
\left[ P( y \neq y^m ) \times \left( 1 - 2 P( y^* \neq y^m ) \right) + P( y^* \neq y^m ) \right] \times (1 - 2 \text{BE})  + \text{BE} = 
\]
\[
\left[ \text{var}(y) \times \left[ 1 - 2 E_{\dataset,t} \left( L\left(  y^*,y^m  \right)   \right) \right] + E_{\dataset,t} \left( L\left(  y^*,y^m  \right)   \right) \right] \times (1 - 2 \text{BE})  + \text{BE} = 
\]
\[
\left[ \text{var}(y) \times \left[ 1 - 2  L\left(  y^*,y^m  \right) \right] + L\left(  y^*,y^m  \right) \right] \times (1 - 2 \text{BE})  + \text{BE} = 
\]
\[
\left[ \text{var}(y) \times \left[ 1 - 2 \times \text{bias}(y^m) \right] + \text{bias}(y^m) \right] \times (1 - 2 \text{BE})  + \text{BE} = 
\]
\[
\text{BE} +  \text{bias}(y^m) (1-2BE) + \text{var}(y) \times \left[ 1 - 2 \times \text{bias}(y^m) \right] \times (1 - 2 \text{BE}) = 
\]
\begin{equation}
\label{BVdecomp_total}
\text{BE} +  \text{bias}(y^m) (1-2BE) + \text{var}(y) \times (1 - 2 \text{BE}) - \left[ 2 \times \text{var}(y) \times \text{bias}(y^m) \right] \times (1 - 2 \text{BE}),
\end{equation}

\noindent where var$(y) \equiv P( y \neq y^m )$, bias$(y^m) \equiv E_{\dataset,t} \left( L\left(  y^*,y^m  \right)   \right) $, and $E_{\dataset,t} \left( L\left(  y^*,y^m  \right)   \right)  =  L\left(  y^*,y^m  \right) $ since $y^*$ and $y^m$ are constants. Note that unlike the case in regression, where the bias and variance enter the bias-variance decomposition in an additive way, the bias-variance decomposition for classification involves an interaction term between the bias and the variance. The role of this term can can be understood as follows. First, note that the Bayes Error cannot be greater than 0.5, and therefore $1 - 2BE \geq 0$. If the bias is zero, that is $y^m = y^*$, then Eq.(\ref{BVdecomp_total}) reduces to

\[
\text{BE} + \text{var}(y) \times (1 - 2 \text{BE}).
\]

\noindent Here, the lower the variance term var$(y)$, the lower the expected prediction error $E_{\dataset,t} \left( L\left(  t,y  \right)   \right)$. However, if the bias is 1, meaning $y^m \neq y^*$, then Eq.(\ref{BVdecomp_total}) will be equal to 

\[
\text{BE} +  (1-2BE) + \text{var}(y) \times (1 - 2 \text{BE}) - 2 \text{var}(y) \times (1 - 2 \text{BE}) =
\]
\[
(1 - \text{BE}) - \text{var}(y) \times (1 - 2 \text{BE}).
\]

\noindent In this case, increasing the variance term var$(y)$ will decrease the expected prediction error $E_{\dataset,t} \left( L\left(  t,y  \right)   \right)$. An increase of  var$(y)$ means that $P(y \neq y^m)$ increases. Since $y^m \neq y^*$ in the biased case, having a $y$ that is different from $y^m$ is a desirable property, which eventually leads to a decrease in the expected prediction error.

We stress that Eq.(\ref{BVdecomp_total}) refers to a given point $\bf{x}$, and is not a bias-variance decomposition over all points in the domain of the data. To arrive at the latter, we need to take the expectation of Eq.(\ref{BVdecomp_total}) w.r.t. all points $\bf{x}$ in the domain of the data. We also note that the bias at point $\bf{x}$ can take one of just two possible values: the bias is equal to 1 if the mean prediction of all $y$ predictions from datasets $\dataset$ coincide with the Bayes prediction at point $\bf{x}$, and it is 0 otherwise. The variance is defined as the proportion of time the predictions $y$ from models built on each dataset $\dataset$ are different from the mean prediction $y^m$ from each of these datasets.

\section{Optimal linear combination of models within the bias-variance framework}
\label{sec:optLinComb}

In the previous section we have decomposed the expected (test) error of a classifier at point $\bf{x}$ to its bias, variance and Bayes-error components (see Eq.(\ref{BVdecomp_total})). Here we will use this bias-variance framework to derive the optimal linear combination of the predictions from different classifiers over a whole dataset of $n$ points. Over all $N$ points from a random training dataset of size $N$ generated by a given underlying function with some noise, the empirical expected prediction error is:

\begin{equation}
\label{dataset_Eerror}
\frac{1}{N} \sum_{i=1}^{N}E_{\dataset,t} \left( L\left(  t_i,y_i  \right)   \right) = 
 \frac{1}{N} \sum_{i=1}^{N}P(t_i \neq y_i) \approx \frac{1}{N} \sum_{i=1}^{N} \sum_{d=1}^{D}\frac{I \left( t_{i} \neq y_{i}^{d} \right) }{D},
\end{equation}

\noindent where we approximate the expected error over $D$ number of datasets, $t_i$ is the realized class label of the $i^{th}$ point in the empirical dataset at hand, $y_{i}^{d}$ is the prediction for $t_i$ from model built on dataset $d$, and $I(\cdots)$ is a zero-one indicator function. We note that the Bayes prediction for point $i$, $y_i^*$, would be 1 if the expected value of $t_i >=0.5$ over infinitely many datasets, and 0 otherwise, even though in the empirical setting we observe one class realization ($t_i$) for the $i^{th}$ point\footnote{The assumption that one point should be observed once in the training empirical dataset is not material, as we treat each point in this dataset separately.}. In case the predicted label $y_i$ originates from a linear combination of $K$ classifiers, and we redefine the classes to be in the set $\{-1,+1\}$, then Eq.(\ref{dataset_Eerror}) will be expanded as:

\begin{equation}
\label{dataset_Eerror_comb}
\frac{1}{N} \sum_{i=1}^{N} \sum_{d=1}^{D}\frac{I \left( t_{i} \neq y_{i}^d \right)}{D} = 
\frac{1}{N} \sum_{i=1}^{N} \sum_{d=1}^{D}\frac{I \left( t_{i} \neq  sign \left( \sum_{k=1}^{K}y_i^{kd} w_k \right)   \right) }{D},
\end{equation}

\noindent where $y_i^{kd}$ is the prediction label for the $i^{th}$ point of the $k^{th}$ classification model built on dataset $d$, $w_k$ is the weight assigned to the $k^{th}$ model, and  $y_i^d =  sign(\sum_{k=1}^{K}y_i^{kd} w_k)$ is the predicted classification label $\in \{-1,+1\}$ of the linear combination of $K$ classifiers on dataset $d$ for point $i$. Under the assumption that each of the $K$ models has expected accuracy over the data of more than 50\% we would require that $1 \geq w_k \geq 0$ and $\sum_{k=1}^{K}w_k = 1$. A classification error at the $i^{th}$ instance in dataset $d$ will occur if $t_i \neq sign(\sum_{k=1}^{K}y_i^{kd} w_k)$, or 

\[
t_i \times sign(\sum_{k=1}^{K}y_i^{kd} w_k) = -1
\]
\[
\Longleftrightarrow
\]
\[
t_i \left(  \sum_{k=1}^{K}y_i^{kd} w_k  \right) < 0 .
\]

\noindent Therefore we would require to have weights $w_k, k=1..K$, such that for each instance $i$ it holds that $t_i \left(  \sum_{k=1}^{K}y_i^{kd} w_k  \right) \geq 0$. If that is not possible, then we introduce a non-negative slack variable, $\xi_{id}$, which would be positive in case this inequality does not hold. Thus, we end up with a constraint for each point $i$ in dataset $d$:
\[
t_i \left(  \sum_{k=1}^{K}y_i^{kd} w_k  \right) \geq 0 - \xi_{id} , ~ ~ ~ i = 1..N, ~ d = 1..D.
\]

\noindent Note that $\sum_{k=1}^{K}y_i^{kd} w_k$ lies in the interval $\left[ -1 , +1 \right]$. We introduce one more requirement, which is that the predictions $y_i^{kd}$ are out-of-sample, so that we keep the independence assumption between them and corresponding instances $t_i$. Possible ways to achieve this is to perform a leave-one-out prediction scheme for each instance, a cross-validation scheme where all predictions are taken from cross-validation test folds (and models are built on training cross-validation folds), or a bootstrap scheme, where we make of use many datasets and make sure that the predicted instances are always out-of-sample for each bootstrap training dataset. The summation over $D$ in Eq.(\ref{dataset_Eerror_comb}) points towards the bootstrap scheme being the closest approximation. 

We can now formulate an optimization problem to find optimal weights $w_k, k=1..K$, for a (meta) classifier, which is a linear combination of $K$ classification models:

\begin{equation}
\label{optimization_comb}
\min_{\mathbf{w,\xi}} \sum_{d=1}^D \sum_{i=1}^N \xi_{id}
\end{equation}
\[
s.t.
\]
\[
t_i \left(  \sum_{k=1}^{K}y_i^{kd} w_k  \right) \geq 0.5 - \xi_{id} , ~ ~ ~ i = 1 .. N, ~ d= 1 .. D
\]
\[
\xi_{id} \geq 0, ~ ~ ~  i = 1..N, ~ d= 1 .. D
\]
\[
1 \geq w_k \geq 0, ~ ~ ~  k = 1..K,
\]
\[
\sum_{k=1}^K w_k = 1,
\]

\noindent where $t_i \in \{ -1 , +1 \}$, $d$ is a bootstrap sample (of a total of $D$ such samples), and $y_i^{kd} \in \{ -1 , +1 \}$ is an out-of-sample $k^{th}$-model prediction for instance $i$ from dataset $d$ in which instance $i$ has been removed. In addition, we have imposed constraints $t_i \left(  \sum_{k=1}^{K}y_i^{kd} w_k  \right) \geq 0.5$ rather than $t_i \left(  \sum_{k=1}^{K}y_i^{kd} w_k  \right) \geq 0$, because predicting class ``0'' should be considered a mistake and associated with a positive $\xi_{id}$ in the optimum.

Note that one option to consider in order to simplify Eq.(\ref{optimization_comb}) is to group constrains for each $d$ (for a given $i$) together. If we sum them up and take the average (dividing by $D$), we end up with corresponding constraints:

\[
\frac{1}{D}\sum_{d=1}^D \left( t_i \left(  \sum_{k=1}^{K}y_i^{kd} w_k  \right) \right) \geq 0.5 - \frac{1}{D} \sum_{d=1}^D \left( \xi_{id} \right) , ~ ~ ~ i = 1 .. N
\]
\[
\Longleftrightarrow
\]
\[
\frac{1}{D} t_i \left( \sum_{d=1}^D \sum_{k=1}^{K}y_i^{kd} w_k   \right) \geq 0.5 - \frac{1}{D} \sum_{d=1}^D \left( \xi_{id} \right) , ~ ~ ~ i = 1 .. N
\]
\[
\Longleftrightarrow
\]
\[
t_i \left( w_1 \frac{1}{D}\sum_{d=1}^D y_i^{1d} + w_2 \frac{1}{D}\sum_{d=1}^D y_i^{2d} + \cdots + w_K \frac{1}{D}\sum_{d=1}^D y_i^{Kd}   \right) \geq 0.5 - \frac{1}{D} \sum_{d=1}^D \left( \xi_{id} \right) , ~ ~ ~ i = 1 .. N
\]
\[
\Longleftrightarrow
\]
\[
t_i \left( w_1 \left( P(y_i^1 = +1) - P(y_i^1 = -1) \right) + \cdots  )   \right) \geq 0.5 - \frac{1}{D} \sum_{d=1}^D \left( \xi_{id} \right) , ~ ~ ~ i = 1 .. N
\]
\[
\Longleftrightarrow
\]
\[
t_i \left( w_1 \left( 2 P(y_i^1 = +1) - 1 \right) + \cdots  )   \right) \geq 0.5 - \frac{1}{D} \sum_{d=1}^D \left( \xi_{id} \right) , ~ ~ ~ i = 1 .. N
\]
\[
\Longleftrightarrow
\]
\[
t_i \left( \sum_{k=1}^K w_k \left( 2 P(y_i^k = +1) - 1 \right) \right) \geq 0.5 - \xi_{i}, ~ ~ ~ i = 1 .. N,
\]

\noindent where we have defined $\xi_{i} \equiv (1/D)\sum_{d=1}^D \left( \xi_{id} \right)$. Therefore, instead of optimization problem Eq.(\ref{optimization_comb}), we have:


\begin{equation}
\label{optimization_comb_prob}
\min_{\mathbf{w,\xi}} \sum_{i=1}^N \xi_{i}
\end{equation}
\[
s.t.
\]
\[
t_i \left(  \sum_{k=1}^{K} \left( 2 P(y_i^k = +1) - 1 \right) w_k  \right) \geq 0.5 - \xi_{i} , ~ ~ ~ i = 1 .. N,
\]
\[
\xi_{i} \geq 0, ~ ~ ~  i = 1..N,
\]
\[
w_k \geq 0, ~ ~ ~  k = 1..K,
\]
\[
\sum_{k=1}^K w_k = 1,
\]

\noindent where the constraints $1 \geq w_k, ~ k=1..K,$ have been dropped, as they are redundant\footnote{One may also consider to drop the constraint $\sum_{k=1}^K w_k = 1$}. Thus, instead of having $D$ (bootstrap) datasets with estimated out-of-sample labels $y_{i}^{kd}$ to optimize over, we can have a cross-validation scheme, where at each point $i$ (in a given test fold) for each model $k$ we have an estimate for $P(y_i^k = +1)$, which we cast as $2P(y_i^k = +1) - 1$ to accommodate for our choice of labeling being $\{-1,+1\}$ rather than $\{0,1\}$. The posterior probability at point $i$, $P(y_i^k = +1)$, can be estimated using, e.g., raw-score calibration via Isotonic regression or implied posterior probability estimation as in \citep{nalbantov2019note}. An even more simplified optimization scheme would be to use, for each point $i$, the raw classification score from the training dataset in a cross-validation mode, instead of $P(y_i^k = +1)$. In this case the raw score for a point $i$ is obtained from a cross-validation test fold, which contains $i$. We follow such a cross-validation scheme in the experimental section \ref{sec:Illustration} below. This will be an adequate alternative to using posterior probability estimates if the raw score and the posterior probability estimates are linearly related.


It is possible to formulate a penalized version of Eq.(\ref{optimization_comb_prob}), where the coefficients $w_k, k=1..K$ are penalized (shrunk) towards $1/K$, corresponding to assignment of equal weight to each classifier. In Eq.(\ref{optimization_comb_prob}) this can be achieved by using constraints for the allowed values of $w_k$ to be $C \geq w_k \geq 0$, where $C$ is a manually adjustable parameter. If $C = 1/K$ the resulting weights are all equal, and if $C \geq 1$ the resulting weights are the same as without modification of the original formulation. 

A possible quadratic penalization version the problem of finding an optimal linear combination of classifiers is:

\begin{equation}
\label{optimization_comb_prob_penalized}
\min_{\mathbf{w,\xi}} ~ ~ \frac{1}{2}\sum_{k=1}^K w_k^2 +  C \sum_{i=1}^N \xi_{i}
\end{equation}
\[
s.t.
\]
\[
t_i \left(  \sum_{k=1}^{K} \left( 2 P(y_i^k = +1) - 1 \right) w_k  \right) \geq 1 - \xi_{i} , ~ ~ ~ i = 1 .. N,
\]
\[
\xi_{i} \geq 0, ~ ~ ~  i = 1..N,
\]
\[
w_k \geq 0, ~ ~ ~  k = 1..K,
\]
\[
\sum_{k=1}^K w_k = 1,
\]

\noindent where $C>0$ is a manually-controlled parameter for the trade-off between sum of errors ($\sum_{i=1}^N \xi_i $) and length of $\bf{w}$ vector, $\sum_{k=1}^K w_k^2$. This length is minimal when $w_k = 1/K$, which is the optimal solution when $C$ approaches zero. Thus, the smaller the $C$, the bigger the penalization towards equal weights for the models. Optimization formulation (\ref{optimization_comb_prob_penalized}) resembles the SVM classifier optimization problem, with two additional constraints: $w_k \geq 0, k=1..K$, and $\sum_{k=1}^K w_k = 1$.

\section{Illustration on UCI datasets}
\label{sec:Illustration}

We illustrate the proposed method of optimal linear classifier combination for the binary task on the so-called heart-statlog and german-credit (numerical) datasets from UCI repository \citep{UCI_Dua:2019}. For this purpose we have created for each dataset $K = 156$ Support Vector Machine models with RBF kernels with varying parameters for the $C$ and $g$ parameters (refer to \citep{Burges98atutorial} for an excellent tutorial on SVMs). The range of $C$ parameters was $2^{\{-2:1:10\}}$, and the range for $g$ parameters was $2^{\{-17:1:6\}}$, yielding a grid of a total of $13 \times 12 = 156$ models. The task is to combine them linearly into one overall model following Eq.(\ref{optimization_comb}), resulting in a weighted average of the 156 models, as the weights on the models are constrained to be positive and sum to one. All results were obtained using 5-fold cross-validation. The solution for the weight vector $\bf{w}$ is sparse, which might not be surprising as the objective function in optimization problem (\ref{optimization_comb}) uses the $L_1$ norm over $\xi_i$'s. 

It should be stressed that the weights assigned to the models within the optimal classifiers combination are estimated from an in-sample procedure, and not validated externally. Therefore, the results are prone to the problem of overfitting. 

\medskip

\begin{figure*}[h]
 	\center
 	\hspace*{-2cm}
		\hbox{\hspace{0.9cm} \includegraphics[height=13cm]{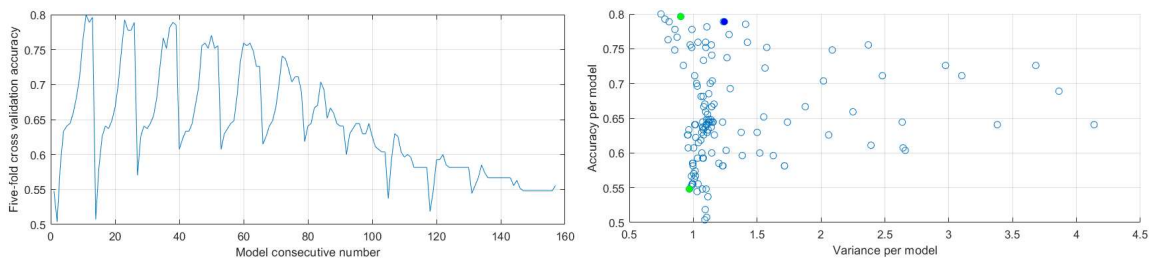}}
 	\vspace*{-9cm}
	\caption{Five-fold cross-validation accuracy for SVM models built using a grid over parameters $C \in 2^{\{-2:1:10\}}$ and $g \in 2^{\{-17:1:6\}}$ (left) and variance-accuracy plot (right). The variance here is calculated as the variance of the differences between raw predicted scores and actual (numeric) labels. Filled circles correspond to models withs weights $\geq 0.1$ (green) and weights $\in (0,0.1)$ (blue) in the optimal linear combination. Dataset: heart-statlog.}
	\label{fig:cvacc_portfolio_heart}
 \end{figure*}

\begin{figure*}[h]
 	\center
 	\hspace*{-2cm}
		\hbox{\hspace{0.9cm} \includegraphics[height=13.2cm]{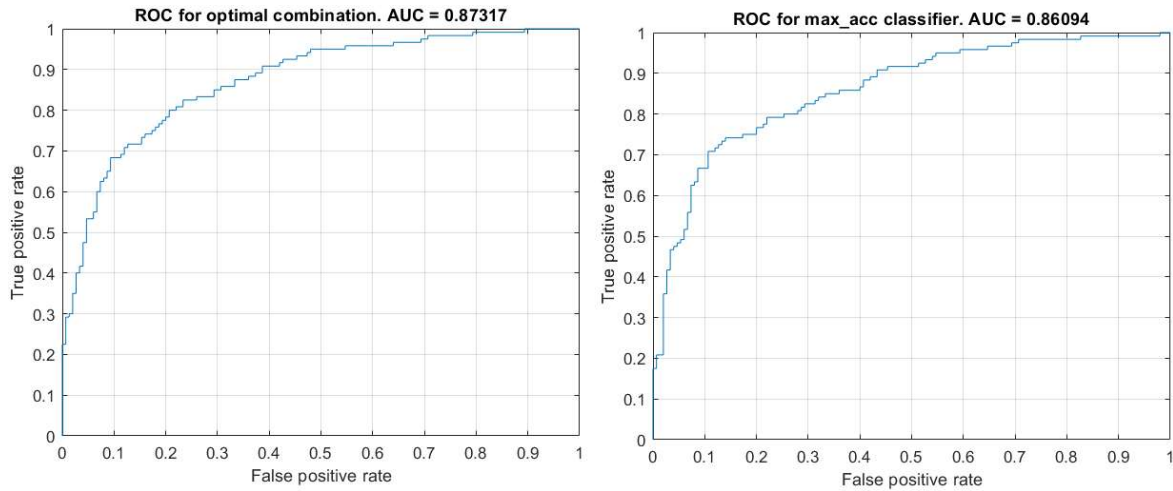}}
 	\vspace*{-6cm}
	\caption{ROC curves and respective AUC's for the optimal classifier combination (left) and the maximal-accuracy classifier (right). The ROC curves are obtained from 5-fold cross validation.  Dataset: heart-statlog.}
	\label{fig:auc_opt_max_heart}
 \end{figure*}

\begin{figure*}[h]
 	\center
 	\hspace*{-2cm}
		\hbox{\hspace{0.9cm} \includegraphics[height=13.2cm]{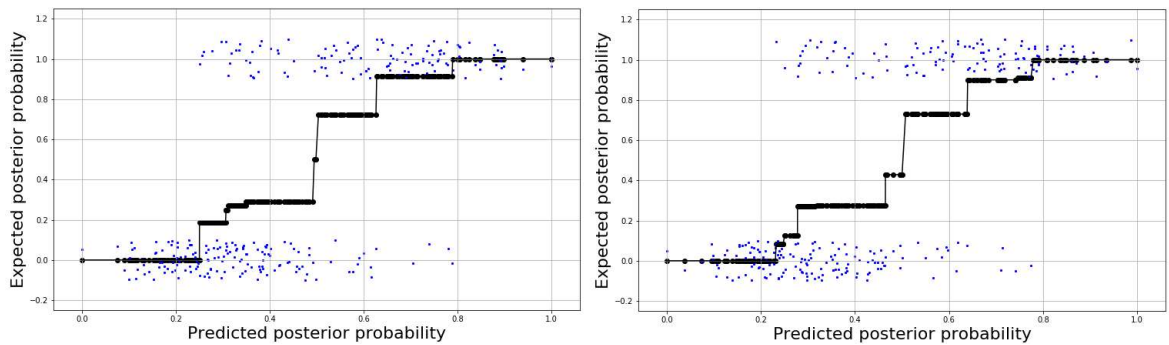}}
 	\vspace*{-8.5cm}
	\caption{Isotonic regression for optimal linear combination of classifies (left) and maximal-accuracy classifier (right). Observed labels are jittered around values 0 and 1 for better visibility. Dataset: heart-statlog.}
	\label{fig:isotonic_opt_max_heart}
 \end{figure*}

Results for the heart-statlog dataset:

\medskip

\noindent The 5-fold cross-validation accuracies for all 156 models are shown in Figure \ref{fig:cvacc_portfolio_heart}, left panel. Out of 156 models, only three weights are positive (the rest being zeros): for model with $C=1024$ and $g=7.62e-06$ which has accuracy of 79.63\% (the weight being $0.5375$), for model with $C=1024$ and $g=1.53e-05$ which has accuracy of 78.89\% (the weight being $0.0635$), and model with $C=1024$ and $g=0.0156$ which has accuracy of 54.81\% (the weight being $0.3990$). The models have consecutive numbers 13, 26 and 156, respectively. We note that the model with the highest cross-validation accuracy (80\%) has received a weight of zero. In Figure \ref{fig:cvacc_portfolio_heart}, right panel, we show a scatterplot between so-called ``variance per model'', defined as the variance of the differences between raw predicted scores and actual (numeric) labels, and accuracy per model. The (cross-validation) AUC's for both classification approaches is similar: $0.87317$ for the optimal linear classifier combination approach, and $0.86094$ for the maximal-accuracy classifier approach (see Figure \ref{fig:auc_opt_max_heart}). We also show the calibration results from isotonic regression estimation for the optimal linear combination classifier as well as for the maximal-accuracy classifier in Figure \ref{fig:isotonic_opt_max_heart}. The mean absolute error between the expected and predicted (via the the isotonic regression) posterior probability is 0.143 for the optimal linear classifier combination and 0.138 for the maximal-accuracy classifier. 

\medskip

\begin{figure*}[h]
 	\center
 	\hspace*{-2cm}
		\hbox{\hspace{0.9cm} \includegraphics[height=13cm]{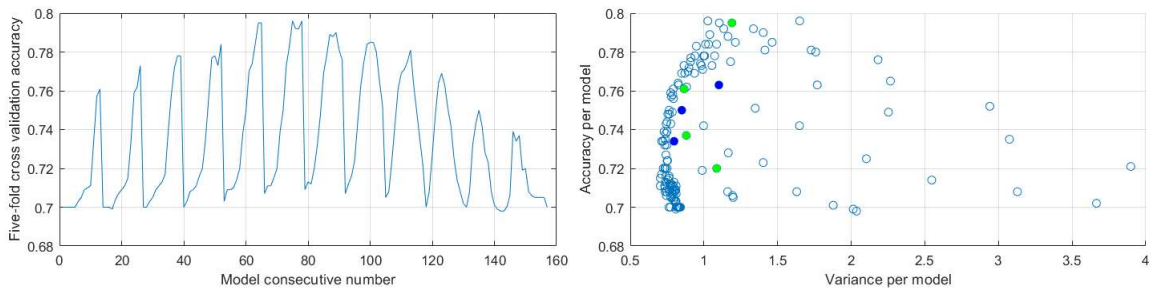}}
 	\vspace*{-8cm}
	\caption{Five-fold cross-validation accuracy for SVM models built using a grid over parameters $C \in 2^{\{-2:1:10\}}$ and $g \in 2^{\{-17:1:6\}}$ (left) and variance-accuracy plot (right). The variance here is calculated as the variance of the differences between raw predicted scores and actual (numeric) labels. Filled circles correspond to models withs weights $\geq 0.1$ (green) and weights $\in (0,0.1)$ (blue) in the optimal linear combination. Dataset: german-credit (numeric).}
	\label{fig:cvacc_portfolio_german}
 \end{figure*}

\begin{figure*}[h]
 	\center
 	\hspace*{-2cm}
		\hbox{\hspace{0.9cm} \includegraphics[height=13.2cm]{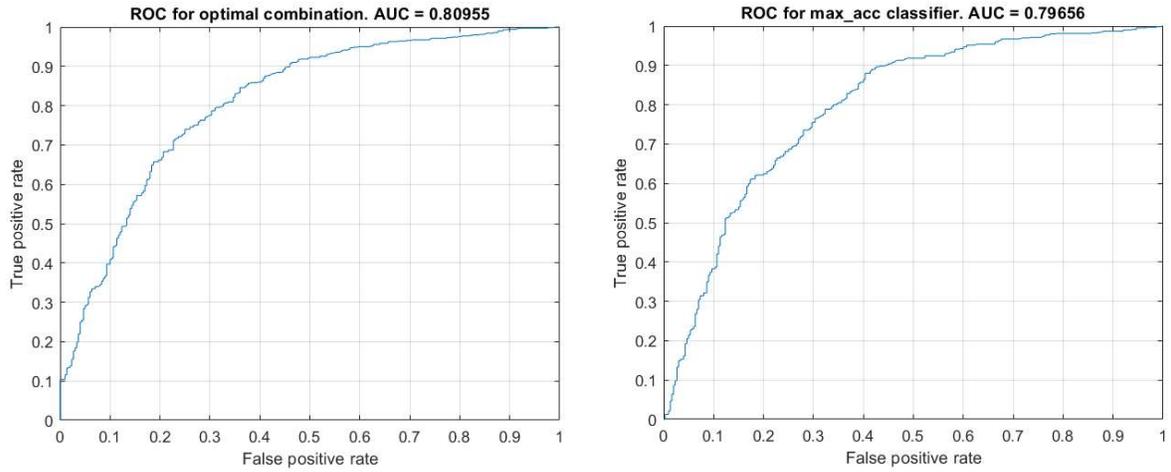}}
 	\vspace*{-6.5cm}
	\caption{ROC curves and respective AUC's for the optimal classifier combination (left) and the maximal-accuracy classifier (right). The ROC curves are obtained from 5-fold cross validation. Dataset: german-credit (numeric).}
	\label{fig:auc_opt_max_german}
 \end{figure*}

\begin{figure*}[h]
 	\center
 	\hspace*{-2cm}
		\hbox{\hspace{0.9cm} \includegraphics[height=13.2cm]{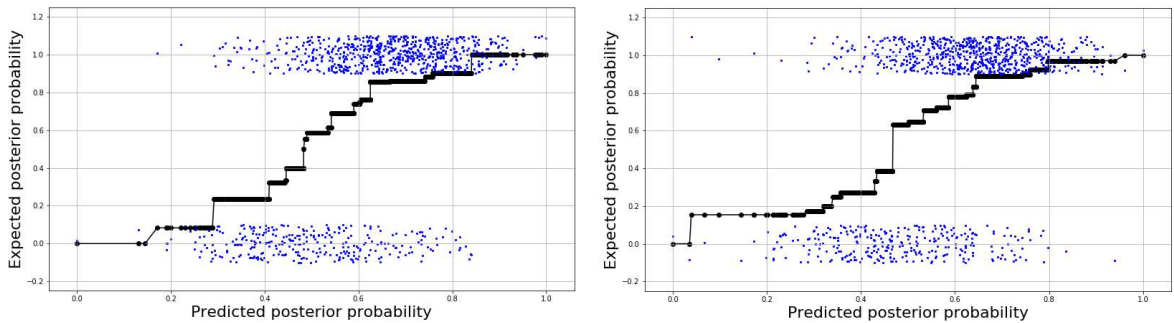}}
 	\vspace*{-8.5cm}
	\caption{Isotonic regression for optimal linear combination of classifies (left) and maximal-accuracy classifier (right). Observed labels are jittered around values 0 and 1 for better visibility. Dataset: german-credit (numeric).}
	\label{fig:isotonic_opt_max_german}
 \end{figure*}

Results for the german-credit (numeric) dataset:

\medskip

\noindent The 5-fold cross-validation accuracies for all 156 models are shown in Figure \ref{fig:cvacc_portfolio_german}, left panel. Out of 156 models, only 7 weights are positive (the rest being zeros).  The models have consecutive numbers 13, 65, 124, 135, 147, 148, and 150, respectively. The corresponding cross-validation accuracies are 76.1\%, 79.5\%, 76.3\%, 75\%, 73.4\%, 73.7\% and 72\%, with model weights, 0.2847, 0.2835, 0.0092, 0.0867, 0.0335, 0.18, and 0.1224, respectively. We note that the model with highest cross-validation accuracy (79.6\%) has received a weight of zero. In Figure \ref{fig:cvacc_portfolio_german}, right panel, we show a scatterplot between``variance per model'' and accuracy per model. The (cross-validation) AUC's for both classification approaches is similar: $0.80955$ for the optimal linear classifier combination approach, and $0.79656$ for the maximal-accuracy classifier approach (see Figure \ref{fig:auc_opt_max_german}). We also show the calibration results from isotonic regression estimation for the optimal linear combination classifier as well as for the maximal-accuracy classifier in Figure \ref{fig:isotonic_opt_max_german}. The mean absolute error between the expected and predicted (via the the isotonic regression) posterior probability is 0.13 for the optimal linear classifier combination and 0.15 for the maximal-accuracy classifier.

\section{Discussion}
\label{sec:discussion}

In this section we discuss in further detail the fundamental block behind the optimal combination of classifiers, namely the bias-variance framework for the classification task. In Section \ref{sec:BV} we have discussed the case when the predicted points are out-of-sample, which makes them independent of the points in the dataset used for training. He we will detail the difference in the estimation of expected prediction error in case of a lack and presence of such dependence. The difference in the estimation is the amount of overfitting incurred by using the training set to evaluate the performance of a model built on this training set rather than evaluate the performance on a test set. We start with reiterating the general bias-variance decomposition of Eq.(\ref{BVdecomp_long}):

\[
E_{\dataset,t} \left( L\left(  t,y  \right)   \right)  = P(t \neq y^{*} \mid y=y^{*}) P(y=y^{*}) +  P(t = y^{*} \mid y \neq y^{*}) P(y \neq y^{*}),
\]

\noindent which holds irrespective of whether $y$ and $t$ are independent or not. In case training data is used to predict $t$, where $t$ is also from the training data, then the prediction $y$ and the training datapoint $t$ are dependent. In this case we cannot drop the ``given'' condition in the above equation and proceed as in Eq.(\ref{BVdecomp_ind}). In case of dependency between $y$ and $t$, Eq.(\ref{BVdecomp_long}) will be expanded as \footnote{We have used the equality $P(A\mid B)P(B) = P(B\mid A)P(A)$}$~^{,}$\footnote{We have used the equality $P(A \mid \sim B) = \frac{P(\sim B \mid A) P(A)}{P(\sim B)} = \frac{(1 - P(B \mid A)) P(A)}{1-P(B)}   = \frac{P(A) - P(A \mid B)P(B)}{1 - P(B)}$}:

\[
E_{\dataset,t} \left( L\left(  t,y  \right)   \right)  = P(y \neq y^{*} \mid t=y^{*}) P(t=y^{*}) +  P(y = y^{*} \mid t \neq y^{*}) P(t \neq y^{*}) = 
\]
\[
\left( 1 - E_{\dataset,t} \left( L\left(  t,y^*  \right)   \right) \right) P(y \neq y^{*} \mid t=y^{*})  + E_{\dataset,t} \left( L\left(  t,y^*  \right)   \right) P(y = y^{*} \mid t \neq y^{*}) = 
\]
\[
= E_{\dataset,t} \left( L\left(  t,y^*  \right)   \right) \left[ P(y = y^{*} \mid t \neq y^{*})  - P(y \neq y^{*} \mid t = y^{*})   \right] + P(y \neq y^{*} \mid t = y^{*}) =
\]
\[
= \text{BE} \left[ 1 - P(y \neq y^{*} \mid t \neq y^{*})  - P(y \neq y^{*} \mid t = y^{*})   \right] + P(y \neq y^{*} \mid t = y^{*}) =
\]
\[
= \text{BE} - \text{BE} \times P(y \neq y^{*} \mid t \neq y^{*})  + P(y \neq y^{*} \mid t = y^{*})(1 - \text{BE}) =
\]
\[
= \text{BE} - \text{BE} \times P(y \neq y^{*} \mid t \neq y^{*})  + \frac{P(y \neq y^{*})  - P(y \neq y^{*} \mid t \neq y^{*}) P(t \neq y^*)}{1 - P(t \neq y^*)}    (1 - \text{BE}) =
\]
\[
= \text{BE} - \text{BE} \times P(y \neq y^{*} \mid t \neq y^{*})  + \frac{P(y \neq y^{*})  - P(y \neq y^{*} \mid t \neq y^{*}) \times \text{BE}}{1 - \text{BE}}    (1 - \text{BE}) =
\]
\[
= \text{BE} - \text{BE} \times P(y \neq y^{*} \mid t \neq y^{*})  + P(y \neq y^{*})  - P(y \neq y^{*} \mid t \neq y^{*}) \times \text{BE} =
\]
\[
= \text{BE} - 2\text{BE} \times P(y \neq y^{*} \mid t \neq y^{*})  + P(y \neq y^{*}) =
\]
\begin{equation}
\label{BVdecomp_dep}
= \text{BE} - 2\text{BE} \times P(y \neq y^{*} \mid t \neq y^{*})  + E_{\dataset,t} \left( L\left(  y,y^*  \right)   \right)
\end{equation}

Eq.(\ref{BVdecomp_dep}) decomposes the expected prediction error at point $\bf{x}$ in the general case, where $t$ and $y$ are dependent. Thus, the difference in the expected prediction error between the independent, $E_{\dataset,t} \left( L\left(  t,y  \right)   \right)^{\text{ind}}$, and dependent, $E_{\dataset,t} \left( L\left(  t,y  \right)   \right)^{\text{dep}}$, cases is:

\[
E_{\dataset,t} \left( L\left(  t,y  \right)   \right)^{\text{ind}} - E_{\dataset,t} \left( L\left(  t,y  \right)   \right)^{\text{dep}}=
\]
\[
\text{BE} - 2\text{BE} \times P(y \neq y^{*})  + P(y \neq y^{*}) -
\]
\[
\left( \text{BE} - 2\text{BE} \times P(y \neq y^{*} \mid t \neq y^{*})  +P(y \neq y^{*}) \right)
\]
\begin{equation}
\label{ind_minus_dep}
2\text{BE} \times \left( P(y \neq y^{*} \mid t \neq y^{*})   - P(y \neq y^{*}) \right).
\end{equation}

This difference, which may be referred to as ``optimism'', is the amount of overfitting between a training and a test sets (at a point $\bf{x}$). In case of dependency between $t$ and $y$, which occurs when $y$ is used in the estimation of $t$, we expect invariably that this dependency is positive. In other words, $P(y \neq y^{*} \mid t \neq y^{*}) > P(y \neq y^{*})$, and therefore $E_{\dataset,t} \left( L\left(  t,y  \right)   \right)^{\text{ind}} - E_{\dataset,t} \left( L\left(  t,y  \right) \right)^{\text{dep}} > 0$.

Let us illustrate the amount of overfitting for the one-nearest-neighbor (1NN) classifier. For 1NN the prediction $y$ at a point $\bf{x}$ is equal to the observed class $t$ at point $\bf{x}$, that is we have $t = y$, assuming all points in the dataset have different coordinates. Therefore, 

\[
E_{\dataset,t} \left( L\left(  t,y  \right)   \right)^{\text{dep}} =\text{BE} - 2\text{BE} \times P(y \neq y^{*} \mid t \neq y^{*})  +P(y \neq y^{*}) =
\]
\[
\text{BE} - 2\text{BE} \times P(t \neq y^{*} \mid t \neq y^{*})  +P(t \neq y^{*}) =
\]
\[
\text{BE} - 2\text{BE} + \text{BE} = 0.
\]

\noindent Thus, if we evaluate the performance of 1NN over point $\bf{x}$, the training error is 0, which is natural for the 1NN classifier. However, the expected error over a test set (at the same point $\bf{x}$) is:

\[
E_{\dataset,t} \left( L\left(  t,y  \right)   \right)^{\text{ind}} =\text{BE} - 2\text{BE} \times P(y \neq y^{*} )  +P(y \neq y^{*}) =
\]
\[
\text{BE} - 2\text{BE} \times P(t \neq y^{*} )  +P(t \neq y^{*}) =
\]
\[
2\text{BE} - 2\text{BE} ^2.
\]

\noindent Naturally, over out-of-sample sets, we will observe at point $\bf{x}$ values of $t$ either 0's or 1's, where the proportion of 1's will be $P(t = 1 \mid \mathbf{x})$, which is not zero. In a training set, the probability that at point $\bf{x}$ label 1 is observed is $P(t = 1 \mid \mathbf{x})$, therefore the probability that label 0 is observed is $1 - P(t = 1 \mid \mathbf{x})$. Let us assume that $P(t = 1 \mid \mathbf{x}) > 0.5$, in which case the optimal predicted class is 1. Thus, if we have just one training dataset with one observation for point $\bf{x}$, we expect to incur the Bayes error, $P(t = 0 \mid \mathbf{x})$, over future instances at this point if the prediction is 1, which happens with expected value $P(t = 1 \mid \mathbf{x})$. Otherwise, if the prediction is 0, which happens with (the lower) expected value $P(t = 0 \mid \mathbf{x})$, then over future test sets the expected error will be $P(t = 1 \mid \mathbf{x})$. Therefore, the expected prediction error is $P(t = 1 \mid \mathbf{x}) \times P(t = 0 \mid \mathbf{x}) + P(t = 0 \mid \mathbf{x}) \times P(t = 1 \mid \mathbf{x}) $ $= (1-BE)BE + BE(1-BE) = 2BE(1-BE) = 2BE - 2BE^2$, which is exactly the derived expected prediction error in the independence case ($y$ and $t$ are independent). We have just derived the general formula for the $E_{\dataset,t} \left( L\left(  t,y  \right)   \right)^{\text{ind}} $ for the 1NN in a logical way.

\begin{figure*}[h]
 	\center
 	\hspace*{-2cm}
		\hbox{\hspace{4cm} \includegraphics[height=8cm]{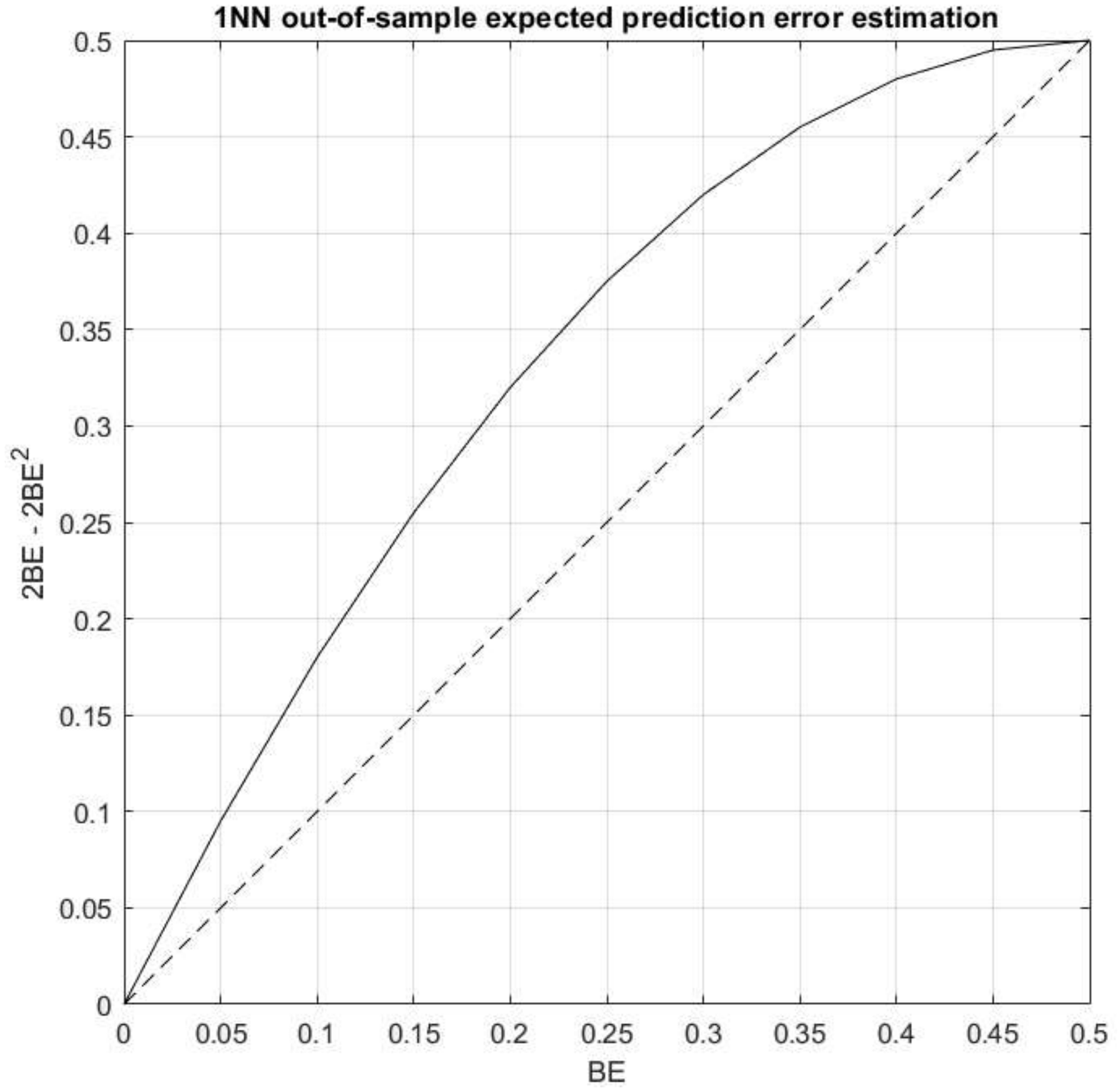}}
 	\vspace*{0cm}
	\caption{The relation between BE and expected out-of-sample prediction error at point $\bf{x}$, $E_{\dataset,t} \left( L\left(  t,y  \right)   \right)^{\text{ind}} = 2\text{BE} - 2\text{BE} ^2$, for the 1NN classifier.}
	\label{fig:1NN_BE}
 \end{figure*}

The relation between BE and $E_{\dataset,t} \left( L\left(  t,y  \right)   \right)^{\text{ind}} = 2\text{BE} - 2\text{BE} ^2$ is shown in Figure \ref{fig:1NN_BE}. Note that the 1NN out-of-sample error is never more than twice the Bayes error, which is a known result. It approaches twice the Bayes error when the BE $\rightarrow$ 0, and is equal to the Bayes error then BE = 0.5.


\section{Conclusion}
\label{conclusion}
We have proposed an approach for the estimation of the optimal linear combination of classifiers (for the binary classification case). The approach has been derived within the bias-variance framework, which we discussed extensively, and illustrated on a dataset from the UCI repository. Further research could center on comparison between this approach and other stacking approaches proposed in the literature.

\vskip 0.2in
\bibliography{ref}

\end{document}